\documentclass[10pt,twocolumn,letterpaper]{article}

\usepackage{iccv}
\usepackage{times}
\usepackage{epsfig}
\usepackage{graphicx}
\usepackage{amsmath}
\usepackage{amssymb}
\usepackage{comment}
\usepackage{xcolor,colortbl}
\usepackage{stfloats}
\usepackage{caption}

\usepackage[pagebackref=true,breaklinks=true,letterpaper=true,colorlinks,bookmarks=false]{hyperref}

\iccvfinalcopy 

\def\iccvPaperID{9152} 

\ificcvfinal\fi

\newcommand{\OURS}{Patch2CAD}

\begin{document}

\title{\OURS: Patchwise Embedding Learning for In-the-Wild Shape Retrieval from a Single Image}


\author{
  Weicheng Kuo\textsuperscript{\rm 1}, Anelia Angelova\textsuperscript{\rm 1}, Tsung-Yi Lin\textsuperscript{\rm 1}, Angela Dai\textsuperscript{\rm 2}\\
  \textsuperscript{\rm 1} Google Research, Brain Team \\
  \textsuperscript{\rm 2} Technical University of Munich \\
  {\tt\small \{weicheng, anelia, tsungyi\}@google.com, angela.dai@tum.de}
}

\twocolumn[{%
	\renewcommand\twocolumn[1][]{#1}%
	\maketitle
	\begin{center}
		\captionsetup{type=figure}
		\vspace{-0.4cm}
		\includegraphics[width=0.95\linewidth]{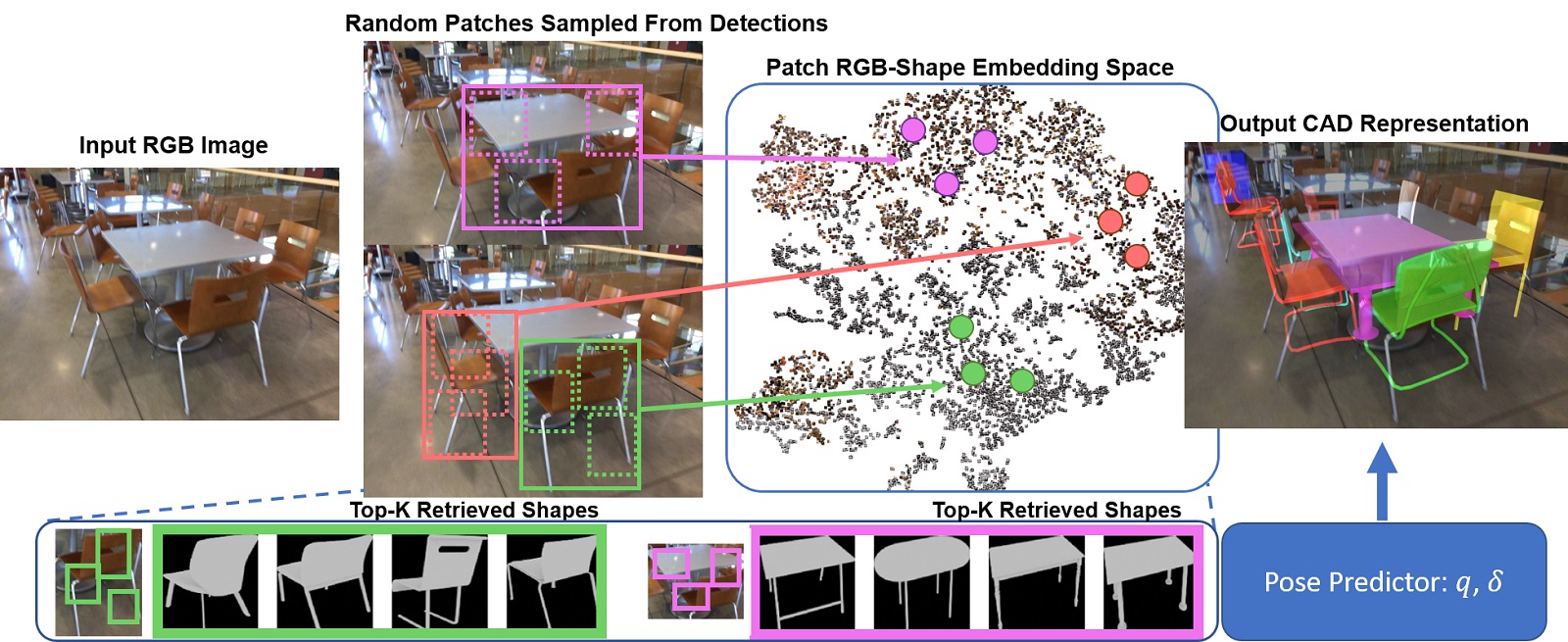}
		\vspace{-0.3cm}
		\captionof{figure}{
		From an input RGB image, we learn a shared image-CAD embedding space by embedding patches of detected objects from the RGB images and patches of CAD models. By establishing patch-wise correspondence between image and CAD, we can establish object correspondence based on part similarities, enabling more effective shape retrieval for new views, as well as robust top-$k$ CAD retrieval. Our patch-based retrieval for a similar 3D CAD representation coupled with pose prediction results in a CAD-based 3D understanding of the objects in the image.
		}
		\label{fig:teaser}
	\end{center}
}]

\maketitle

\newlength\savewidth\newcommand\shline{\noalign{\global\savewidth\arrayrulewidth
  \global\arrayrulewidth 1pt}\hline\noalign{\global\arrayrulewidth\savewidth}}

\newcommand{\tablestyle}[2]{\setlength{\tabcolsep}{#1}\renewcommand{\arraystretch}{#2}\centering\footnotesize}

\newcommand{\ourpar}[1]{\vspace{1mm}\noindent\textbf{#1}}

\newcommand*\rot{\rotatebox{75}}
\definecolor{Gray}{gray}{0.95}
\definecolor{Gray2}{rgb}{0.4, 0.4, 0.4}
\definecolor{LightCyan}{rgb}{0.88,1,1}
\definecolor{Lavender}{rgb}{0.909, 0.909, 0.95}

\newcommand{\Fone}[1]{F1$^{#1}$}

\newcolumntype{a}{>{\columncolor{Lavender}}c}
\newcolumntype{b}{>{\columncolor{white}}c}
\newcolumntype{T}{>{\footnotesize}c} 

\newcommand{\LL}{\mathcal{L}}

\newcommand{\APbox}{AP$^{\text{box}}$~}
\newcommand{\APmask}{AP$^{\text{mask}}$~}
\newcommand{\APmesh}{AP$^{\text{mesh}}$~}

\interfootnotelinepenalty=10000

\begin{abstract}
   3D perception of object shapes from RGB image input is fundamental towards semantic scene understanding, grounding image-based perception in our spatially 3-dimensional real-world environments.
   To achieve a mapping between image views of objects and 3D shapes, we leverage CAD model priors from existing large-scale databases, and propose a novel approach towards constructing a joint embedding space between 2D images and 3D CAD models in a patch-wise fashion -- establishing correspondences between patches of an image view of an object and patches of CAD geometry.
   This enables part similarity reasoning for retrieving similar CADs to a new image view without exact matches in the database. 
   Our patch embedding provides more robust CAD retrieval for shape estimation in our end-to-end estimation of CAD model shape and pose for detected objects in a single input image.
   Experiments on in-the-wild, complex imagery from ScanNet show that our approach is more robust than state of the art in real-world scenarios without any exact CAD matches.
\end{abstract}



\section{Introduction}
Fundamental to many visual perception tasks is an understanding of the decomposition of an observed scene into its constituent objects, and the semantic meaning of these objects -- including class categorization, segmentation, and structural 3D understanding.
In recent years,  advances in 2D object recognition and localization have achieved impressive success in image-based understanding, even from only single image input \cite{he2016deep,girshick2015fast,ren2015faster,he2017mask}.
Such recognition and perception constrained to the image domain unfortunately remains limited towards understanding 3D attributes such as shape and structure, which are not only fundamental towards a comprehensive, human-like understanding of objects in a scene, but towards many applications such as autonomous exploration and interaction of an environment.

To address 3D perception from a single RGB image, we have recently seen several methods proposed taking a generative approach towards reconstructing the observed objects' geometry \cite{meshrcnn, nie2020total3dunderstanding, denninger2020}. 
These methods show promising results in attaining 3D understanding of objects in complex scene imagery, but take a low-level approach towards geometric reconstruction, constructing voxel-by-voxel or vertex-by-vertex, often resulting in noisy or oversmoothed geometry, or geometry not representing a valid object instance (\eg, missing a leg on a chair).
In contrast, several approaches have taken a CAD-based prior for representing the 3D structure of the objects seen in an RGB or RGB-D observation, by retrieving and aligning CAD models from a database similar to the observed objects  \cite{nan2012search,li2015database,Avetisyan2019Scan2CAD,kuo2020mask2cad}.
This CAD prior enables representation of each object with a clean, complete 3D mesh known to represent valid instances of objects and able to be stored compactly for potential downstream applications.
Unfortunately, such a retrieval-based approach tends to struggle with generalization, in particular when a new observed image of an object does not exactly match any CAD model in the dataset.

We observe that in these challenging scenarios, various part similarities can be leveraged to find a similar shape. Thus, we propose \OURS{}, which constructs a joint embedding space between images and CAD models based on encoding mid-level geometric relations by establishing similarity of patches of images to patches of object geometry. 
These correspondences can be aggregated into CAD prediction by majority voting. This enables CAD retrieval based on the predominant part similarities, enabling improved generalizability for CAD retrieval to reconstruct the shapes of objects seen in an image. 

To achieve a 3D understanding of object structure from a single RGB image, we first detect object locations in 2D, then construct our patch-based image-CAD embedding space enabling voting for retrieval of a similar CAD model, and predict the pose of that CAD in the image.
\OURS{} is trained end-to-end to comprehensively establish an effective image-CAD embedding.

Our main contribution is a patch-based learning of a joint embedding space between the two very different domains of 2D images and 3D CAD models, which establishes more robust, part-level-based correspondences (see Figure~\ref{fig:teaser}).
We demonstrate that this patch-wise embedding enables meaningful CAD retrievals for image observations not just in the top nearest neighbor, but for top-$k$ retrieval.
As a result, we achieve more effective association of CAD shapes to images observations of objects with no exact CAD matches in a candidate database, as is typically the case for real-world scenarios. 
We demonstrate \OURS{}'s effective 3D shape perception on both ScanNet \cite{dai2017scannet} and Pix3D \cite{sun2018pix3d} datasets.
In particular, on the complex, in-the-wild images from ScanNet~\cite{dai2017scannet}, \OURS{} exhibits notable advantage to its patch-based approach, outperforming state of the art by 1.9 Mesh AP (22\% relative improvement).

\section{Related Work}

Scene understanding is one of the fundamental problems in computer vision. 
A vast amount of literature on the topic has forwarded the field in understanding of 2D images: for example popular methods for object detection~\cite{girshick2015fast,ren2015faster,redmon2016you,liu2016ssd,lin2017focal,law2018cornernet,duan2019centernet,wu2020rethinking}, semantic segmentation~\cite{long2015fully,huang2017densely} and instance segmentation~\cite{he2017mask,kuo2019shapemask}.
Our approach is inspired by these 2D image understanding approaches, but instead focuses on producing a 3D representation of the objects observed in a single image, providing additional geometric, structural information about the scene.

\vspace{-3mm}
\paragraph{Single-View Object Reconstruction.}
Recently, we have also seen remarkable progress in reconstructing the 3D shape of an object from a single RGB image.
Such work has also been driven by shape representations: earlier research focused on dense volumetric grids~\cite{choy20163d,wu2016learning}, while point clouds~\cite{fan2017point,yang2019pointflow} and hierarchical structures such as octrees~\cite{tatarchenko2017octree,riegler2017octnet} offered more memory and computationally efficient representations.
Mesh-based approaches offer an efficient surface representation along with adaptive structure, but tend to rely on strong topological assumptions, taking a deformation-based approach from a given template mesh~\cite{wang2018pixel2mesh,wen2019pixel2mesh++}; generative approaches without relying on templates tend to be limited to small numbers of vertices \cite{dai2019scan2mesh}.
Implicit functions have recently seen notable success in single-object shape reconstruction, characterizing shape by prediction an occupancy or signed distance field value  for a location in space \cite{mescheder2019occupancy,park2019deepsdf,saito2019pifu}. Approaches to predict convex primitives have also been shown to produce promising results \cite{deng2020cvxnet}.

While these approaches operated on images encompassing only one object,  Mesh R-CNN pioneered an approach for generating the shapes for multiple objects seen in an RGB image, which more closely represents real-world perception scenarios.
Several methods have now furthered development on this task; Mask2CAD~\cite{kuo2020mask2cad}  proposed a CAD retrieval-based approach towards understanding the shape and pose of objects, and Nie \etal~\cite{nie2020total3dunderstanding} a mesh generation approach for object reconstruction based on initial deformation from a sphere followed by edge refinement to handle local topology errors.
Our approach also tackles shape reconstruction for the multiple objects seen in an RGB image, leveraging CAD retrieval and focusing on the construction of a robust image-CAD embedding space.

\paragraph{CAD-Based Retrieval and Alignment.}
An alternative to generative methods for reconstruction is to leverage CAD model priors to represent objects in a scene, and retrieve and align them to achieve a scene reconstruction composed of clean, compact mesh representations of each object.
Early work in computer vision demonstrated the use of existing geometric models as priors \cite{binford1982survey,chin1986model,roberts1963machine}; the current availability of large-scale CAD model datasets (\eg, ShapeNet~\cite{shapenet2015}, Pix3D~\cite{sun2018pix3d}) has revitalized this approach.
Various methods have been introduced for CAD model retrieval and alignment to RGB-D scans~\cite{shao2012interactive,kim2013guided,li2015database,bansal2016marr,Avetisyan2019Scan2CAD,grabner2019location,izadinia2020licp}, including end-to-end learning pipelines~\cite{avetisyan2019end,avetisyan2020scenecad}, as well as CAD alignment to an image assuming that shape is given ~\cite{lim2013parsing,georgakis2019learning,huang2015single}.

From a single image, Aubry et al.~\cite{aubry2014seeing} develop handcrafted HOG-based features to match textured renderings of CAD models to images in order detect chairs; our approach learns to associate CAD patches with image patches based on a more general learning of purely geometric correspondence, enabling learning geometric structures as well as the use of geometry-only CAD databases.
More recently, Izadinia and Seitz~\cite{izadinia2017im2cad} and Huang \etal~\cite{huang2018holistic} apply analysis-by-synthesis approaches for CAD model alignment and  scene layout estimation from a single image, leveraging a costly optimization (minutes to hours) for each input image.

Shape retrieval methods also show promising results by learning joint RGB-CAD space embedding~\cite{tatarchenko2019single,li2015joint,massa2016deep,kuo2020mask2cad}. 
Li \etal~\cite{li2015joint} propose a method to construct a joint embedding space between RGB images and CAD models, enabling CAD model retrieval from images; the embedding space is first constructed from shape descriptors, and then image embeddings are optimized for into the shape space. Massa \etal~\cite{massa2016deep} learns to adapt object RGB features to CAD space with a projection layer for object instance detection. Kuo \etal~\cite{kuo2020mask2cad} jointly optimize for a shared embedding space between image views and CAD models in order to perform retrieval for multiple objects seen in an image.
Such techniques can be prone to overfitting, as an object shape obtains a single, global representation, and a new image may not contain exact CAD matches but rather various part similarities.
Our approach addresses a similar problem statement in learning a mapping between images and CAD models; however, to better generalize to new observations with inexact matches, leveraging a majority part similarity to more robustly retrieve CAD models for reconstruction. 
\section{Method}

\begin{figure*}
\begin{center}
	\includegraphics[width=0.9\linewidth]{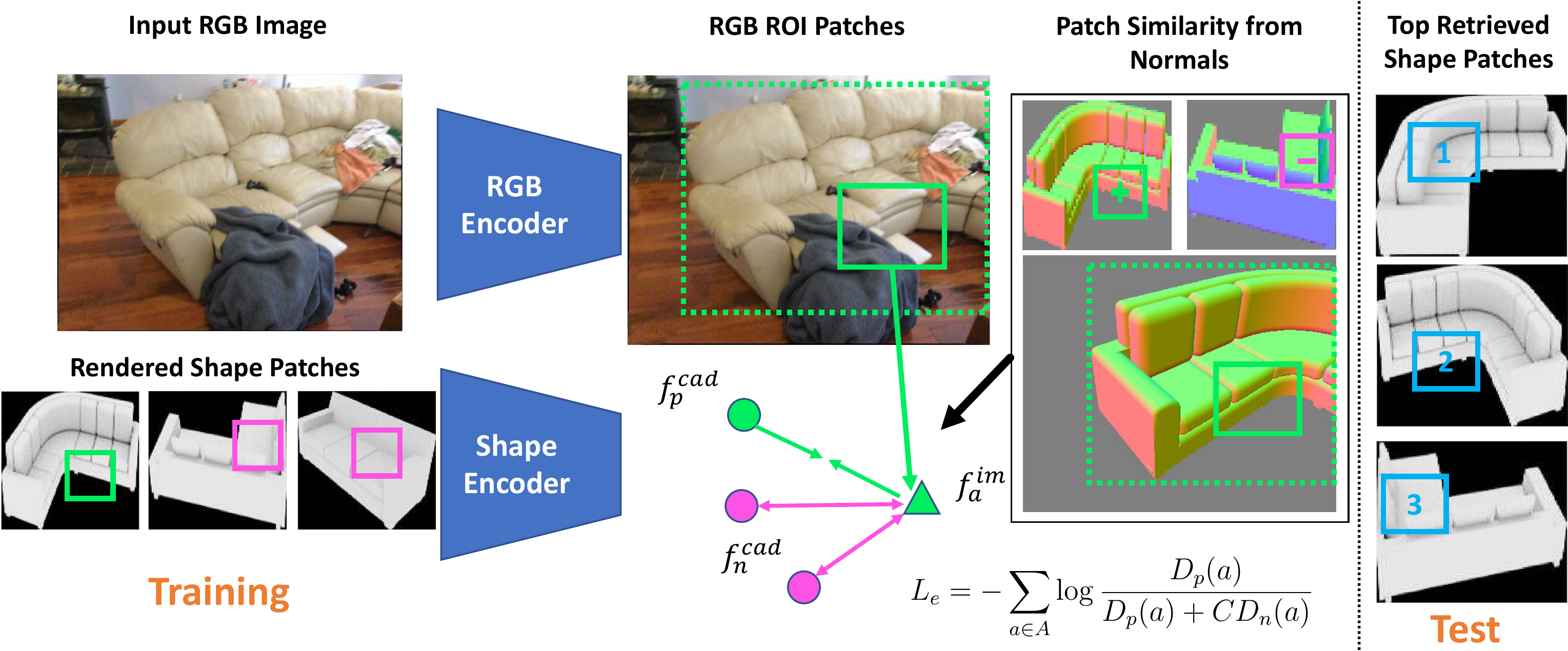}
	\vspace{-0.4cm}
   \caption{Our goal is to learn a shape embedding space for retrieval by leveraging patch correspondence between RGB and shape. At train time, we sample RGB patches from object regions and rendered shape patches from the object class. We shape the embedding space with a contrastive loss, and regularize the learning with surface normal matching so that positive patches have high geometric similarity, while negative patches come from non-matching shapes with low geometric patch similarity.  This patch-wise construction establishes more robust correspondence for shape retrieval from images at test time.}
	\vspace{-0.5cm}
\label{fig:overview}
\end{center}
\end{figure*}

\subsection{Overview}
From a single input RGB image, we aim to understand the observed scene by predicting the object semantics and 3D structures, by retrieving and aligning similar CAD models to the observed image.
Objects are first detected in the 2D image, represented by their 2D bounding box, class label, and 2D instance segmentation mask.
We then aim to learn a shared embedding space between the image representation of the objects and CAD models, in order to retrieve a similar CAD model representing the 3D structure of a detected object.
In a separate pose-prediction head, we simultaneously regress the pose of the CAD model that aligns it with the image observation.

A shared embedding space between image and CAD can be difficult to effectively construct due to the strong differences between the two domains.
While mapping image observations of an object together with the full CAD models into a shared embedding space has shown promise~\cite{li2015joint,kuo2020mask2cad}, this approach tends to struggle with generalization to views of objects without exact matches in the CAD database.
Thus, rather than constructing an embedding space which maps similar image observations of an object together with full CAD models, we aim to learn an embedding space which captures not only global semantic similarity between image and CAD, but mid-level and low-level geometric similarities.
We propose to learn the embedding of object parts and CAD parts by constructing a shared feature space where patches of object images lie close to similar patches of CAD objects.
This enables reasoning about similar parts for retrieval in a scenario without exact CAD matches for a new image view, enabling more robust CAD reconstruction.

\subsection{2D Object Detection}
We leverage a state-of-the-art 2D object detection and instance segmentation backbone to inform our 3D shape reasoning.
From an input RGB image, 2D object bounding boxes and class labels are localized using RetinaNet~\cite{lin2017focal}, and instance segmentation masks are predicted using ShapeMask~\cite{kuo2019shapemask}.
The learned features from the 2D object detection guide our shape prediction; for a detected object $k$, we use the predicted box for an object to crop features $f_k$ corresponding to the object, and multiply with the instance mask prediction $m_k$.
We then use $m_k\circ f_k$ as input for the image-shape embedding as well as pose alignment.

\subsection{Patch-Based Joint Embedding Learning}
\label{sec:embed-learn}
Our approach centers around constructing a patch-based joint embedding space between the two domains of image observations of an object and 3D CAD model representations of objects.
While humans can establish perceptual correspondence between images and CAD models, it is challenging to bridge the domains due to strong differences in representation: in contrast to a 3D geometric CAD model, an image is view-dependent, colored, and contains lighting and material effects.
Moreover, in real-world scenarios, we typically do not have exact CAD matches to the image views as groundtruth annotation. We construct an embedding space between image patches and CAD patches, to enable reasoning about mid-level and low-level structural similarities between objects, as many objects can share similarly structured parts while not matching exactly. 
Thus, we can establish part similarities where a global object mapping can struggle to fulfill a complete object match.
By bridging the two domains in such a fashion, we can more easily recognize similar geometric structures in a new observation of an object not exactly represented in the CAD database.

Our patch-based joint image-shape embedding space is visualized in Figure~\ref{fig:overview}, and is constructed based on patches of the image feature of an object and patches of rendered CAD models to $n$ canonical views $\{c_{0}, ...c_{n-1}\}$ (similar to a light-field descriptor~\cite{chen2003visual}).
The representation of CAD models to their rendered views helps to reduce the domain gap between 2D image and 3D shape; we use $n=16$ views with the canonical views determined by K-medoid clustering of the train views.
To embed image and shape, we extract features from $m_k\circ f_k$ and sampled patches from the rendered CAD views using embedding modules composed of a series of 2D convolutions, resulting in $f^{\textrm{im}}$ and $\{ f^{\textrm{cad}}_{j} \}$, respectively.
Each embedding network for image and CAD features is structured symmetrically, without shared weights as they operate on different domains.
We then construct the patch embedding space by randomly sampling anchor patches from $f^{\textrm{im}}$, which we denote as $f^{\textrm{im}}_a$, and then establish positive and negative similarity with the $\{ f^{\textrm{cad}}_{j} \}$.
In all our experiments, we use a patch size of $1/3$ RGB-ROI or rendered shape image size.
The embedding space is constructed with a noise contrastive estimation loss~\cite{oord2018representation}:
\begin{equation}
    L_e = -\sum_{a\in A} \log \frac{D_p(a)}{D_p(a) + C D_n(a)}
\end{equation}
\begin{equation}
    D_p(a) = \frac{1}{\vert P(a) \vert}\sum_{p\in P(a)}{\exp{(D(f^{\textrm{im}}_{a}, f^{\textrm{cad}}_p))}}
\end{equation}
\begin{equation}
    D_n(a) = \frac{1}{\vert N(a) \vert}\sum_{n\in N(a)}{\exp{(D(f^{\textrm{im}}_{a}, f^{\textrm{cad}}_n))}} 
\end{equation}
where $L_e$ denotes the total loss, $A$ the set of all anchor (query) patches, $P(a)$ and $N(a)$ the positive and negative matches for the query patch $a$, $C=24$ a weighting value, $D(x, y) = (x / ||x||)^T(y / ||y||) / \tau$ with $\tau = 0.15$. $D_p(a)$ and $D_n(a)$ are the mean exponentiated weights of positive and negative pairs. To further improve learning efficiency, we exclude empty RGB and shape patches in the embedding loss, as determined by the rendered binary masks. 

Our overall loss is similar to Mask2CAD, but because we operate on patch-level correspondence, we removed  hard-positive mining due to relaxed constraints on patch matching vs. exact instance matches. The formulation of our loss is different from standard InfoNCE loss because we have multiple positives (shape rendering patch) for each query patch. Thus, we need to balance the ratio of positives/negatives per batch via the $C$ parameter.

\paragraph{Patch similarity for embedding construction.}
To train our embedding construction, we establish patch similarity for matching and non-matching patches between image and CAD patches by estimating their geometric similarity.
We use rendered normals from the CAD models (with normals represented in canonical space) and their corresponding patches to represent local geometry, and for the image, we use patches of the rendered normals of the ground truth corresponding CAD model.
For each patch of normals, we compute its descriptor by a self-similarity histogram, evaluated as the histogram over all pairwise angular distances of the normals in the patch; histograms are normalized to sum to $1$.
This allows us to estimate orientation-independent geometric similarity.
We then measure the difference between two patches of normals by IoU of their self-similarity histograms.
Positive matches to a query are determined by patches corresponding to a ground truth CAD annotation with self-similarity  IoU $> \theta_p$, and negative matches by patches corresponding to non-corresponding CAD models with self-similarity  IoU $< \theta_n$. 
Since a ground truth CAD annotation may contain patches that are dissimilar to the query patch and non-ground truth CAD models may contain patches that are similar to the query, we empirically  found that double thresholds helped to avoid such associations.
We set $\theta_p, \theta_n = 0.4, 0.6$ in our implementation.

We additionally employ hard negative mining by sampling the top negative patches by distance to the query.
During training, we take $16\times$ the number of objects per image for hard negative examples.
This enables for better distinguishing on difficult cases, and an improved embedding space. We set $\vert N(a) \vert = 1024$ for each anchor patch. Regarding hard-positive mining, we observe that it hurts the performance with a fixed top-$K$ mining due to unstable number of positive pairs per batch. To remedy this, we average the weights of all positive pairs and treat them as one positive sample $D_p(a)$, which leads to significantly more stable learning and better performance.

\subsection{Patch-Based Retrieval}
Since our joint embedding of images and shapes is constructed patch-wise, we can leverage many patch retrievals for a more robust, comprehensive shape retrieval.
We use randomly sampled patches from CAD renderings to construct the database for retrieval.
Then for a detected object in an image, we randomly sample $K_q$ patches from $f^{\textrm{im}}$, and for each patch, we retrieve $K_r$ patches from the database.
The $K_r$ retrieved patches are then used to decide the corresponding CAD model of the patch query by majority voting, resulting in $K_q$ CAD models for each patch; the final shape retrieval is obtained by majority vote of the $K_q$ patch-retrieved CADs, excluding those retrieved by patches fully outside the predicted instance mask. While an image-CAD mapping based on full image view of the object and whole CAD model might struggle to retrieve from a global similarity perspective under inexact matches, our patch-based shape retrieval encourages the retrieved shapes to more comprehensively match the image.

\subsection{Pose Prediction}
We simultaneously predict the pose of the 3D shape corresponding to its 2D image observation in a separate branch.
Similar to \cite{kuo2020mask2cad}, we predict the rotation of the shape by a rotation classification followed by a regression refinement, and the translation as an offset from the 2D bounding box center.
To obtain the estimated rotation, use rotation bins computed by K-medoid clustering of the train object rotations as quaternions, and predict the bin using a cross entropy loss, followed by predicting a refinement offset quaternion with a Huber loss \cite{huber1992robust}.
Translation is estimated as an offset from the predicted bounding box center as a ratio of the box dimensions, and optimized with a Huber loss.

\subsection{Implementation Details}

Our ShapeMask~\cite{kuo2019shapemask} instance segmentation backbone (ResNet-50-FPN) is initialized with COCO pretraining, and our embedding for both image and CAD renderings uses a ResNet-18-FPN backbone with random intialization. We train our instance segmentation for amodal bounding box prediction instead of modal boxes in the standard COCO setup, as this can capture more consistent context and provides more stable guidance for the pose translation estimation.
We additionally  apply data augmentation to improve generalization, including HSV color jitter, ROI box jitter, and image scale jitter during training.

We train our approach for 36K iterations using a batch size of 256 on ScanNet, which takes $\approx$ 2 days. The learning rate is initialized to 0.16 and decreased by 10x at 24K and another 10x at 30K iterations. In terms of inference time, Patch2CAD takes $\approx$ 74ms per image \footnote{Measured on Pix3D for comparison with Mask2CAD \cite{kuo2020mask2cad}.}, 58ms model + 16ms retrieval (vs. Mask2CAD $\approx$ 60ms), with unoptimized parallel patch retrievals.

\begin{table*}[tp]
  \centering
  \tablestyle{1.5pt}{1.5}
  \vspace{-0.2cm}
  \begin{tabular}{l|ccc|cccccccc} 
     ScanNet 25K & \cellcolor{blue!10}AP & AP50 & AP75 & \emph{bed} & \emph{sofa} & \emph{chair} & \emph{cabinet}  & \emph{trashbin} & \emph{display} & \emph{table} &\emph{bookshelf}  \\ 
     \shline
     Mask2CAD \cite{kuo2020mask2cad} & \cellcolor{blue!10}8.4 & 23.1 & 4.9 & 14.2 & {\bf 13.0} & 13.2 & {\bf 7.5} & 7.8 & 5.9 & 2.9 & 3.1 \\
     \OURS{} (Ours) & \cellcolor{blue!10}{\bf 10.3} & {\bf 26.0} & {\bf 6.6} & {\bf 18.8} & 12.4 & {\bf 17.6} & {\bf 7.5} & {\bf 8.6} & {\bf 10.8} & {\bf 3.3} & {\bf 3.3} \\
      \hline
  \end{tabular}
  \vspace{-0.2cm}
  \caption{Performance on ScanNet ~\cite{dai2017scannet}. 
        We report mean AP\textsuperscript{mesh} and per-category AP\textsuperscript{mesh}.
  }
  \vspace{-0.2cm}
  \label{table:scannet-main}
\end{table*}

\begin{figure*}
\begin{center}
	\includegraphics[width=0.80\linewidth]{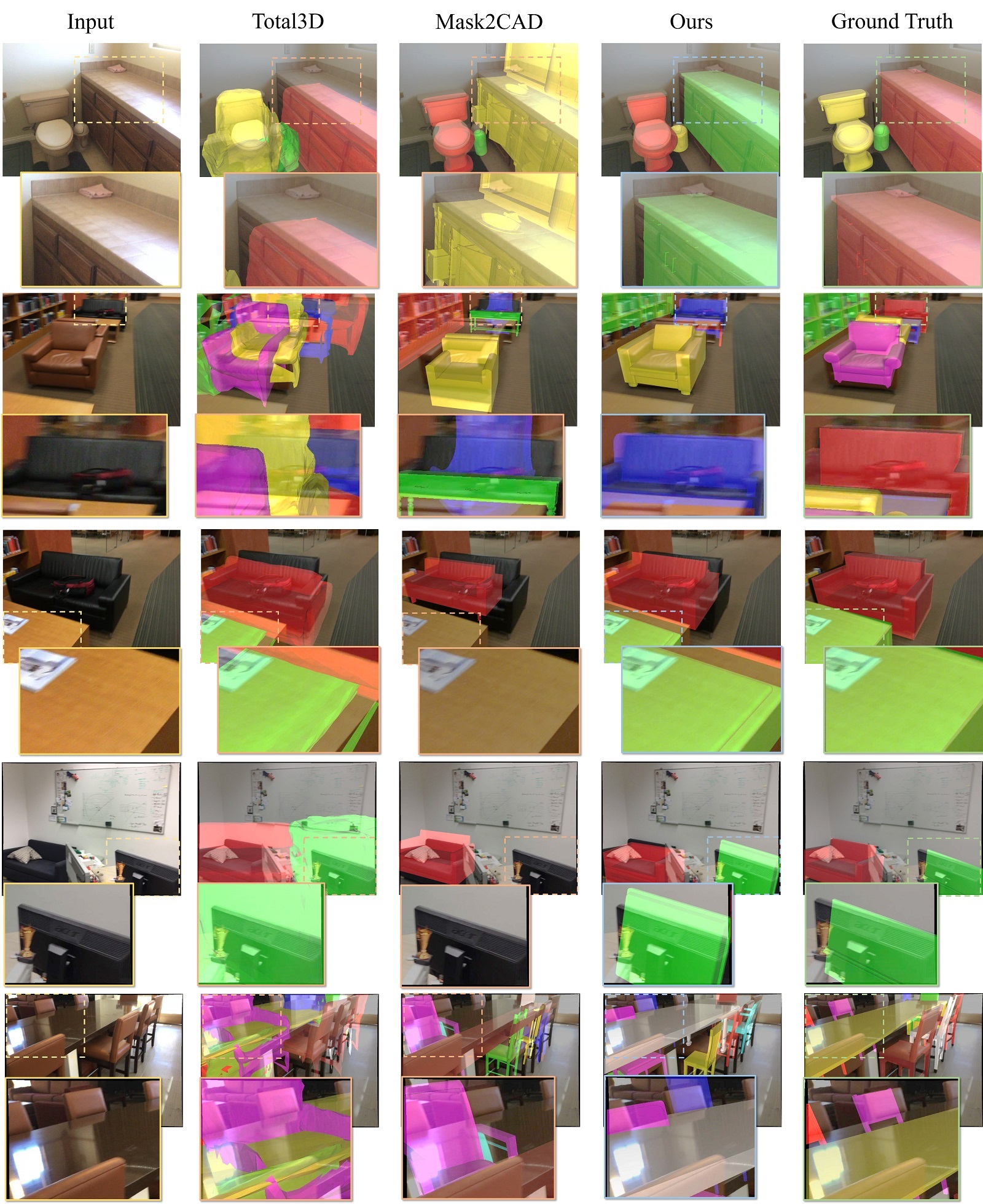}
	\vspace{-0.3cm}
   \caption{Qualitative results on ScanNet~\cite{dai2017scannet} images, in comparison with state of the art Total3D \cite{nie2020total3dunderstanding} and Mask2CAD \cite{kuo2020mask2cad}. Our patch-based shape embedding results in more accurate shape retrieval as well as more robust retrieval for strongly occluded objects (see rows 3, 4, 9, 10). Note that different colors denote distinct object instances in the visualization.}
\label{fig:scannet_comparison}
\end{center}
\end{figure*}

\begin{table*}
  \centering
  \tablestyle{1.5pt}{1.5}
  \vspace{-0.2cm}
  \begin{tabular}{l|ccc|cccccccc} 
     ScanNet 25K & \cellcolor{blue!10}AP & AP50 & AP75 & \emph{bed} & \emph{sofa} & \emph{chair} & \emph{cabinet}  & \emph{trashbin} & \emph{display} & \emph{table} &\emph{bookshelf}  \\ 
     \shline
     Total3D \cite{nie2020total3dunderstanding} & \cellcolor{blue!10} 1.4 & 6.3 & 0.2 & 1.9 & 4.3 & 1.5 & 0.8 & 0.1 & 0.0 & 0.7 & 2.1  \\
     Mask2CAD \cite{kuo2020mask2cad} & \cellcolor{blue!10} 10.5 & 33.3 & 4.5 & 13.9 & {\bf 13.1} & 14.8 & 11.6 & 10.8 & 8.8 & 4.1 & 7.4 \\
     \OURS{} (Ours) & \cellcolor{blue!10} {\bf 12.9} & {\bf 37.5} & {\bf 6.6} & {\bf 14.5} & 11.6 & {\bf 18.8} & {\bf 12.4} & {\bf 13.0} & {\bf 19.0} & {\bf 5.7} & {\bf 8.1}\\
      \hline
  \end{tabular}
  \vspace{-0.2cm}
  \caption{Performance on ScanNet ~\cite{dai2017scannet} using groundtruth 2D detections. 
        We report mean AP\textsuperscript{mesh} and per-category AP\textsuperscript{mesh}.
  }
 \label{table:scannet-gt}
\end{table*}

\begin{table}[h]
\centering
\resizebox{0.45\textwidth}{!}{
\begin{tabular}{| c | c | c | c | c |} 
 \hline
 Method & Patch Size & Normals & AP & AP50 \\ 
 \hline \hline
 Mask2CAD \cite{kuo2020mask2cad} & 1.0 &  & 8.4 & 23.1 \\ 
 \hline
 Patch2CAD (Ours) & 1.0 & V & 9.4 & 24.7 \\ 
 \hline
 Patch2CAD (Ours) & 0.5 & V & 9.5 & 24.5 \\
 \hline
 Patch2CAD (Ours) & \textbf{0.33} & V & \textbf{10.3} & \textbf{26.0} \\
 \hline
 Patch2CAD (Ours) & 0.25 & V & 10.0 & 25.8  \\
 \hline 
\end{tabular}
}
\vspace{-0.2cm}
\caption{Performance vs patch size and use of normals. 1.0 patch size corresponds to the full object size.}
\vspace{-0.4cm}
\label{tab:crop_size}
\end{table}
\begin{table*}
  \centering
  \tablestyle{1.5pt}{1.5}
  \begin{tabular}{l|c|cccccccc} 
     ScanNet 25K & \cellcolor{blue!10}Mean & \emph{bed} & \emph{sofa} & \emph{chair} & \emph{cabinet}  & \emph{trashbin} & \emph{display} & \emph{table} &\emph{bookshelf}  \\ 
     \shline
     Total3D \cite{nie2020total3dunderstanding} & \cellcolor{blue!10} 52.4 & 58.8 & {\bf 72.6} & {\bf 69} & 41.5 & 38.9 & 35.9 & 44.4 & 58.4 \\
     Mask2CAD \cite{kuo2020mask2cad} & \cellcolor{blue!10} 60.6 & 63.1 & 64.4 & 66.1 & {\bf 61.0} & 68.3 & 58.7 & 47.1 & 56.3  \\
     \OURS{} (Ours) & \cellcolor{blue!10} {\bf 63.8} & {\bf 64.3} & 62.0 & 68.1 & 59.9 & {\bf 71.6} & {\bf 73.9} & {\bf 51.9} & {\bf 58.9}  \\
      \hline
    \end{tabular}
  \vspace{-0.2cm}
  \caption{Mean F-score and category F-score on ScanNet ~\cite{dai2017scannet} using groundtruth 2D detections and evaluating shape only. 
  }
  \label{table:scannet-shape-only}
\end{table*}
\begin{table}
  \centering
  \begin{tabular}{l|ccc} 
     Pix3D $\mathcal{S}_1$ & \cellcolor{blue!10}AP & AP50 & AP75 \\ 
     \shline                                       
      Mesh R-CNN \cite{meshrcnn} & \cellcolor{blue!10}17.2 & 51.2 & 7.4 \\
      Mask2CAD \cite{kuo2020mask2cad} & \cellcolor{blue!10}\textbf{33.2} & \textbf{54.9} & \textbf{30.8} \\
      \OURS{} & \cellcolor{blue!10}30.9 & 51.7 & 28.2 \\
      \hline
  \end{tabular}
  \vspace{-0.3cm}
  \caption{Performance on Pix3D~\cite{sun2018pix3d} $\mathcal{S}_1$. We report mean AP\textsuperscript{mesh} following \cite{meshrcnn,kuo2020mask2cad}.
  }
  \label{table:pix3d}
  \vspace{-4mm}
\end{table}


\section{Experiments}

\begin{figure*}
\begin{center}
	\includegraphics[width=0.92 \linewidth]{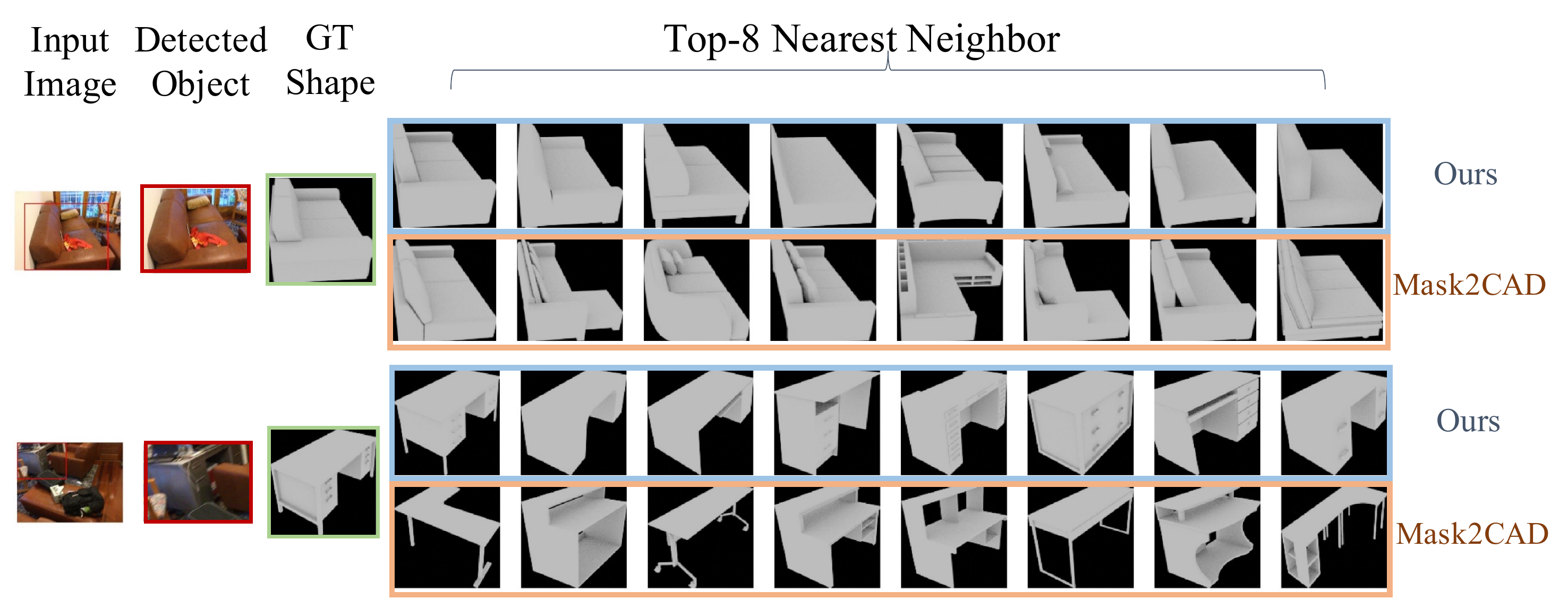}
	\vspace{-0.4cm}
   \caption{Top-$k$ nearest neighbor retrievals from detected objects in ScanNet~\cite{dai2017scannet} images with Scan2CAD~\cite{Avetisyan2019Scan2CAD} ground truth CADs, in comparison to Mask2CAD~\cite{kuo2020mask2cad} (same detection inputs). Our approach achieves a more consistent shape embedding space, enabling robust top-$k$ retrieval with structurally similar CAD associations for not only the top-$1$ nearest neighbor.}
	\vspace{-0.2cm}
\label{fig:topk_comparison}
\end{center}
\end{figure*}

We evaluate our approach on the ScanNet dataset~\cite{dai2017scannet}, which contains challenging real-world imagery of multiple objects per image in cluttered indoor environments, with many occlusions, partial views and varying lighting conditions.
The ScanNet dataset contains 1513 indoor scenes; we use the Scan2CAD~\cite{Avetisyan2019Scan2CAD} annotations of ShapeNet~\cite{shapenet2015} CAD models to the ScanNet scenes to provide ground truth CAD correspondences for training and evaluation.
Note that there are no exact matches between the CAD models to the real-world imagery, as is reflective of many real-world application scenarios.
Following the Mask2CAD~\cite{kuo2020mask2cad} evaluation protocol on ScanNet, we use the 25K frame subset provided by the dataset for training and validation, containing 19387 train and 5436 validation images, respectively. 

In addition, we evaluate our approach on the Pix3D dataset~\cite{sun2018pix3d}, which contains $10,069$ images of indoor furnitures labeled with corresponding CAD models. We use the train/test split by Mesh R-CNN~\cite{meshrcnn} for direct comparison.

\paragraph{Evaluation metrics.}
We adopt previously established metrics for both 2D and 3D evaluation.
For evaluating 2D outputs, we employ the predominant metrics from 2D object recognition: APbox and APmask on the 2D detections of objects.
We use the newly introduced APmesh metric~\cite{meshrcnn} to evaluate the  3D shape and pose predictions for the 3D objects.
Similar to Mask2CAD~\cite{kuo2020mask2cad}, we evaluate APmesh metrics at IoU 0.5 (AP50) at IoU 0.75 (AP75), as well as AP as the mean over AP50-AP95, using 10 IoU thresholds between 0.5 and 0.95. For more consistent reproducibility, we report our evaluations as an average of $2$ independent runs. The thresholds used in F-scores follow ~\cite{kuo2020mask2cad} on ScanNet and \cite{meshrcnn,kuo2020mask2cad} on Pix3D.

\paragraph{Comparison to the state of the art.}
In Table~\ref{table:scannet-main}, we evaluate our 3D object understanding from a single image in comparison to Mask2CAD~\cite{kuo2020mask2cad} on the ScanNet~\cite{dai2017scannet} benchmark proposed by Mask2CAD. Our improvement on Mesh AP50 is larger than AP75, showing that \OURS{} maintains more robust shape estimate even when it retrieves non-exact matches. Mask2CAD also takes a retrieval-based approach, but maps full image observations of an object to entire CAD models, which tends towards overfitting and  struggles with new test images whose objects do not exactly match the database; our patch-level embedding enables more robust retrieval and alignment by establishing correspondence with similar object parts rather than the more strict requirement of the full object.
Additionally, this can help to retrieve and align objects that are occluded or partially visible in the input image (see Figure~\ref{fig:scannet_comparison}).

\paragraph{What is the effect of patch-wise embedding learning on representing shape geometry?}
In Table~\ref{table:scannet-main}, we see that our patch-based embedding improves a retrieval-based 3D object reconstruction, in contrast to the whole-shape embedding of Mask2CAD.
We additionally evaluate our patch-based embedding for retrieval given ground truth 2D detections in Table~\ref{table:scannet-gt}, showing consistent improvement over both the Mask2CAD retrieval and the mesh generation approach of Nie \etal.~\cite{nie2020total3dunderstanding}.
Note that we use the training scheme of Nie \etal. on SUN RGB-D~\cite{song2015sun}, as they use scene layout information during training (similar to ScanNet, SUN RGB-D is also captured from real indoor scenes with a PrimeSense-based sensor).
Finally, we evaluate Patch2CAD given ground truth 2D detections as well as pose (\ie, evaluating shape only) in comparison with Nie \etal.~\cite{nie2020total3dunderstanding} as well as Mask2CAD~\cite{kuo2020mask2cad} in Table~\ref{table:scannet-shape-only}, using an F-score for shape reconstruction evaluation.
Even with a shape-only prediction, strong occlusions in the image views can be challenging; Patch2CAD maintains more robustness.

\paragraph{Effect of patch size and use of surface normals.}
Table~\ref{tab:crop_size} analyzes various patch sizes and with/without surface normals.
The first row corresponds to Mask2CAD; ours on the fourth row.
Our $\frac{1}{3}$ patch basis and use of normals helps to notably  improve shape retrieval.

\paragraph{Comparison on Pix3D.}
Table~\ref{table:pix3d} shows a comparison with both generative and retrieval methods on Pix3D~\cite{sun2018pix3d}. 
Pix3D presents a scenario with exact shape matches in simpler scenes than ScanNet. 
\OURS{} performs notably better than Mesh R-CNN~\cite{meshrcnn}, and is competitive with Mask2CAD~\cite{kuo2020mask2cad}, whose full object matching approach well-suits the scenario with exact 3D matches.

\paragraph{How does a patch-wise embedding mold the space for top-$k$ retrieval?}
We evaluate our patch-based image-CAD embedding space by analyzing the top-$k$ nearest neighbor CAD models retrieved for a given detected object on the ScanNet validation set.
We visualize the top-8 retrieved CAD models for various image object detections in Figure~\ref{fig:topk_comparison}.
In contrast to the full-shape mapping of image-CAD established by Mask2CAD, our patch-wise embedding construction encourages more similarly structured CAD shapes to be voted for by the patches, resulting in geometrically consistent top-$k$.

\begin{figure}
\includegraphics[width=0.95\linewidth]{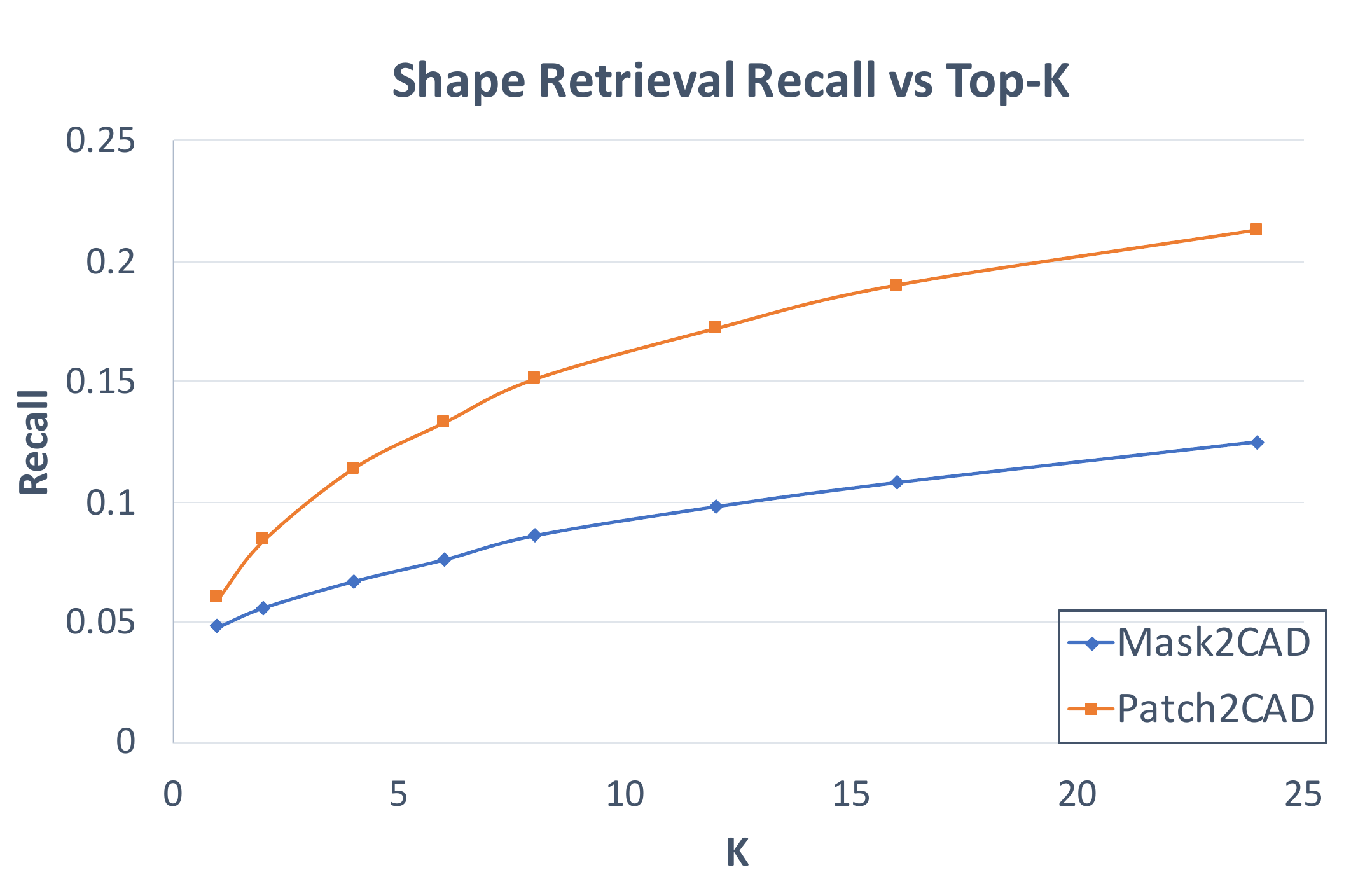}
\vspace{-0.4cm}
   \caption{Shape retrieval comparison with Mask2CAD~\cite{kuo2020mask2cad}}
\label{fig:shape_recall}
\end{figure}

Quantitatively, we analyze our top-$k$ shape retrieval by evaluating the recall from the $k$ retrieved shapes. We compare with the state-of-the-art Mask2CAD~\cite{kuo2020mask2cad} approach in Figure \ref{fig:shape_recall}, using $k=1$ to $24$. Our patch-based approach consistently produces more accurate shape retrieval.

\vspace{-0.4cm}

\vspace{1em}
\noindent \emph{Limitations.}
While our \OURS{} approach demonstrates a more robust joint embedding space construction between images and CAD models, there are various directions for development.
For instance, our patch-based retrieval can produce more robust CAD retrieval results, but cannot represent shapes that differ significantly from the database; we believe a part-based synthesis or deformation approach from our various patch retrievals holds promise.
Additionally, our approach tackles object shape and structure, but does not represent the full scene geometry, which is an important direction towards comprehensive 3D perception.

\section{Conclusion}

In this paper, we present \OURS{}, which establishes patch-based correspondence between 2D images and 3D CAD models for a robust construction of a shared embedding space to map between the two domains.
This enables CAD-based understanding of the shapes of objects seen from a 2D image, representing each object as a posed, lightweight, complete mesh.
We demonstrate that our patch-wise embedding learning can construct a more meaningful embedding space for nearest neighbor retrieval, and more robust shape estimation for complex real-world imagery under many occlusions.
We believe that this brings understanding forward in bridging these domains of 2D-3D as well as real-synthetic, which opens avenues in domain transfer, content creation, and 3D scene understanding.

\section*{Acknowledgements}
We would like to thank our colleagues at Google Research for their advice, and the support of the Bavarian State Ministry of Science and the Arts as coordinated by the Bavarian Research Institute for Digital Transformation (bidt) for Angela Dai.

\clearpage
{\small
\bibliographystyle{ieee_fullname}
\bibliography{egbib}
}

\clearpage






\section*{Appendix A: Additional Top-K Retrieval Qualitative Results}
\begin{figure*}[bp]
\begin{center}
	\includegraphics[width=0.9\linewidth]{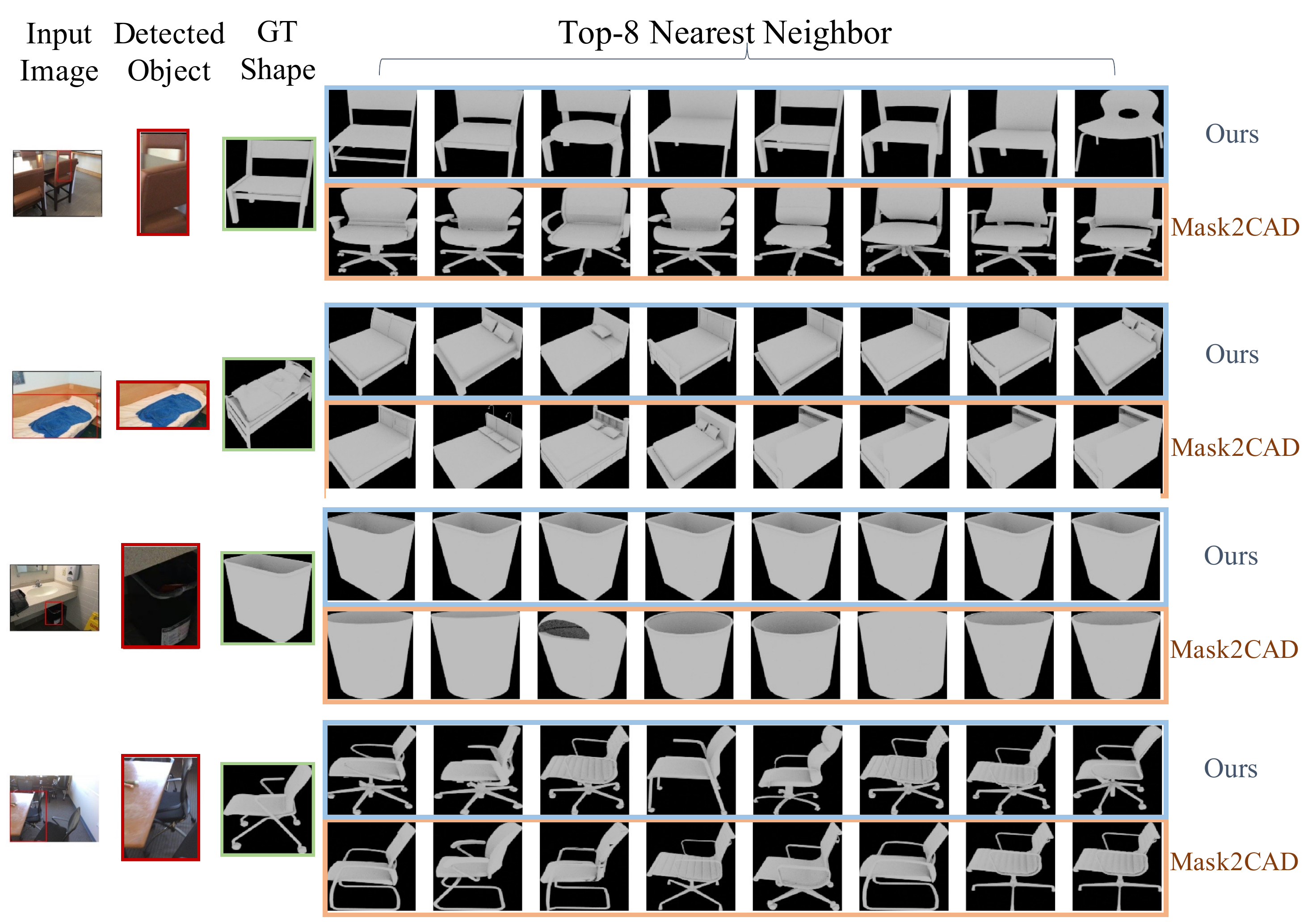}
   \caption{Additional Patch2CAD Top-K retrival qualitative results on various ScanNet~\cite{dai2017scannet} images in comparison with Mask2CAD. }
\label{fig:scannet_top_k}
\end{center}
\end{figure*}

In Figure~\ref{fig:scannet_top_k}, we show additional qualitative results of our Patch2CAD top-K retrieval vs Mask2CAD. We observe that ours can retrieve better and more consistent shapes in the top-K pool than Mask2CAD.

\begin{figure*}
\begin{center}
	\includegraphics[width=0.97\linewidth]{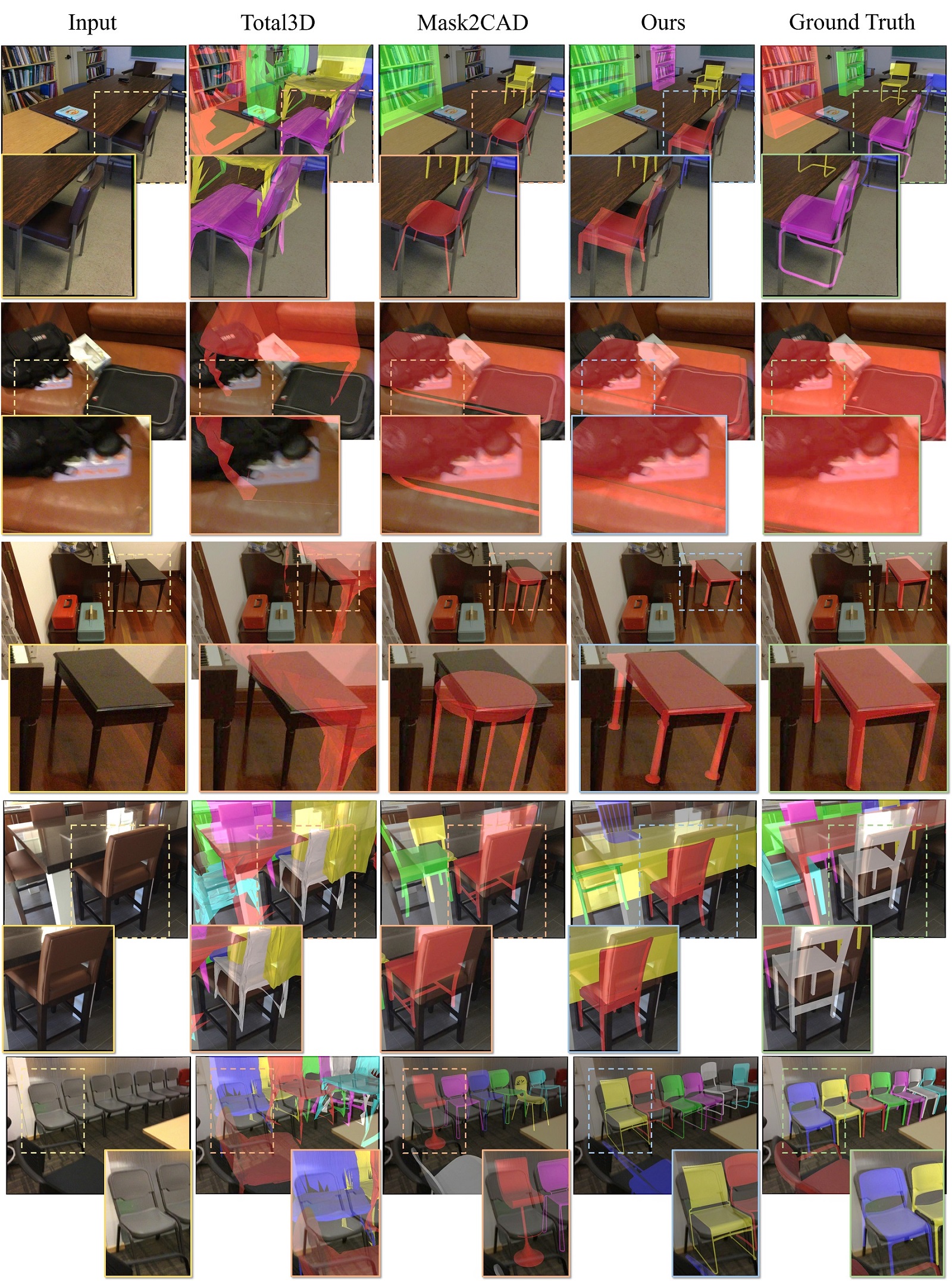}
   \caption{Additional qualitative results of Patch2CAD (ours) on various ScanNet~\cite{dai2017scannet} images. }
\label{fig:scannet_gallery}
\end{center}
\end{figure*}

\section*{Appendix B: Additional Qualitative Results}
In Figure~\ref{fig:scannet_gallery}, we show additional qualitative results of Patch2CAD on ScanNet~\cite{dai2017scannet} images, with Scan2CAD~\cite{Avetisyan2019Scan2CAD} targets. Ours is able to retrieve better matching shapes to the groundtruth than Mask2CAD~\cite{kuo2020mask2cad} or Total3D~\cite{nie2020total3dunderstanding}.

\section*{Appendix C: t-SNE embedding of Patch2CAD}
\begin{figure}
\begin{center}
	\includegraphics[width=\linewidth]{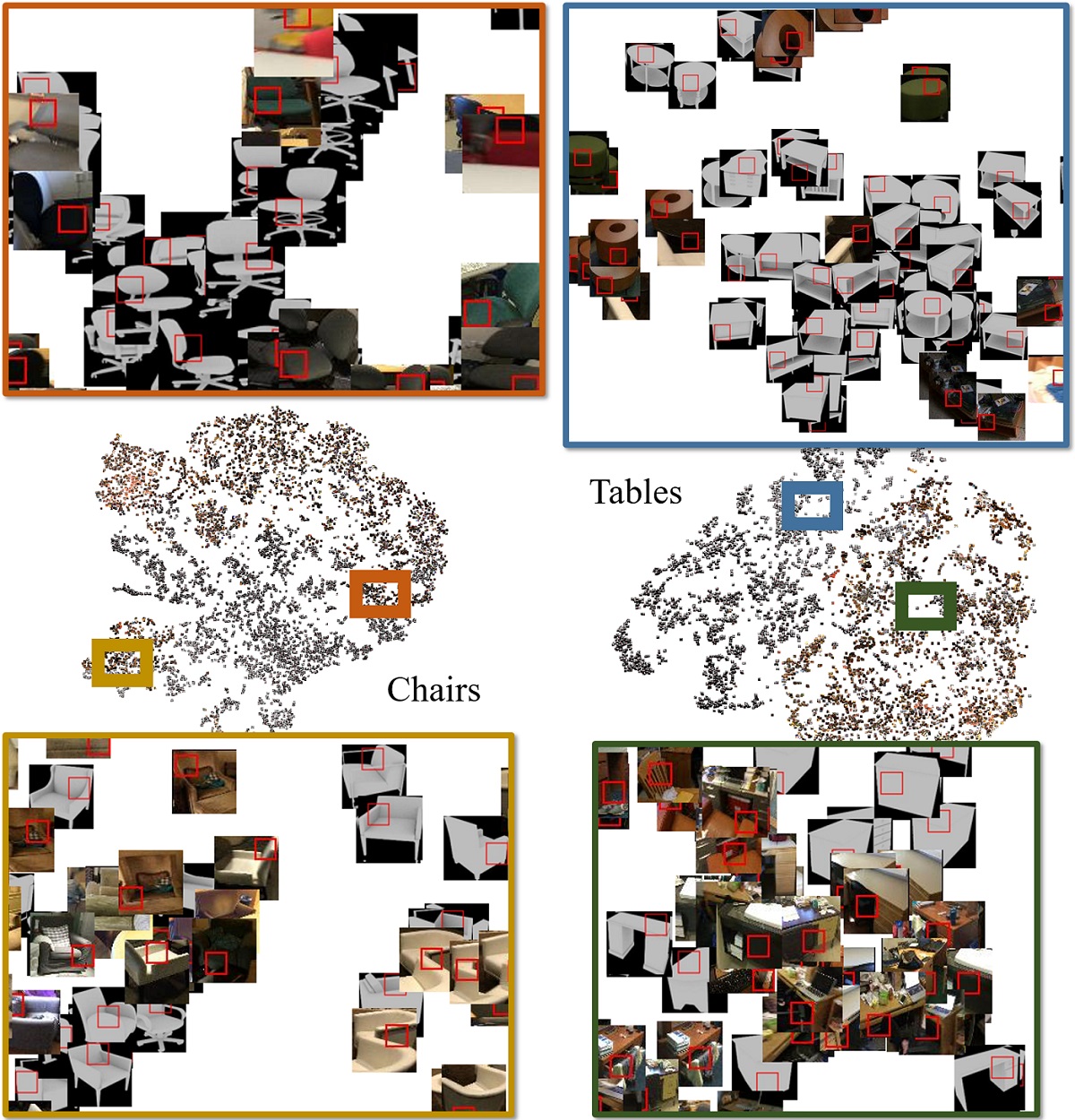}
	\vspace{-0.5cm}
  \caption{t-SNE embeddings of our patch-wise embedding of images and CAD shapes (patches demarcated in red) for the `chairs' and `tables' categories.}
\label{fig:tsne}
\end{center}
\end{figure}

We visualize several t-SNE embeddings in Figure~\ref{fig:tsne}, where CAD patches can tend to cluster near each other (there are many locally very similar patches), but also near similar image patches (\eg, chair seat corner, tabletop).

\section*{Appendix D: Effect of the number of query ($K_q$) and retrieved patches ($K_r$).}
We use one model for all inference time ablation studies in this section. All parameters are the same as the main paper unless stated otherwise. The noise across independent runs are $\approx$ 0.1 Mesh AP.

Table~\ref{tab:kq} analyzes query $K_q$ patches per detection at test time. We see that more patches result in better retrieval.
\begin{table}[h]
\centering
\resizebox{0.32\textwidth}{!}{
\begin{tabular}{| c | c | c | c | c | c |} 
 \hline
 $K_q$ & 1 & 3 & 6 & 9 & 12 \\ 
 \hline \hline
 AP & 9.2 & 9.8 & 10.2 & \textbf{10.3} & 10.2 \\
 \hline 
\end{tabular}
}
\caption{Mesh AP vs the number of query patches.}
\label{tab:kq}
\end{table}

Table~\ref{tab:kr} shows improvement with retrieved $K_r$ per test query, due to robustness of voting when $K_r$ is high.

\begin{table}[h]
\centering
\resizebox{0.4\textwidth}{!}{
\begin{tabular}{| c | c | c | c | c | c | c | c |} 
 \hline
 $K_r$ & 1 & 3 & 6 & 12 & 24 & 48 & 96 \\ 
 \hline \hline
 AP & 9.3 & 9.4 & 9.8 & 10.0 & 10.3 & \textbf{10.6} & 10.6 \\
 \hline 
\end{tabular}
}
\caption{Mesh AP vs the number of retrieved patches.}
\label{tab:kr}
\end{table}

\clearpage

\end{document}